\documentclass[10pt,twocolumn,letterpaper]{article}

\usepackage{iccv}
\usepackage{times}
\usepackage{epsfig}
\usepackage{graphicx}
\usepackage{amsmath}
\usepackage{amssymb}
\usepackage{multirow}
\usepackage{listings}
\usepackage{color}
\usepackage{balance}
\usepackage{caption}
\usepackage{subcaption}
\usepackage{pdfpages}

\newcommand{\mc}[2]{\multicolumn{#1}{c|}{#2}}
\DeclareCaptionFormat{myformat}{#1#2#3\hrulefill}
\lstdefinestyle{customc}{
  belowcaptionskip=1\baselineskip,
  breaklines=true,
  frame=L,
  xleftmargin=\parindent,
  language=C,
  showstringspaces=false,
  basicstyle=\footnotesize\ttfamily,
  keywordstyle=\bfseries\color{green!40!black},
  commentstyle=\itshape\color{purple!40!black},
  identifierstyle=\color{blue},
  stringstyle=\color{orange},
}

\usepackage[pagebackref=true,breaklinks=true,letterpaper=true,colorlinks,bookmarks=false]{hyperref}

\iccvfinalcopy  

\def\iccvPaperID{3933} 

\begin{document}
\title{X-Section: Cross-Section Prediction for Enhanced RGB-D Fusion}

\author{{Andrea Nicastro\textsuperscript{1},  Ronald Clark\textsuperscript{1}, Stefan Leutenegger\textsuperscript{2} }\\
{\textsuperscript{1}Dyson Robotics Lab,  \textsuperscript{2}Smart Robotics Lab,  Imperial College London }\\
{\tt\small a.nicastro15@imperial.ac.uk}
}

\twocolumn[{%
\renewcommand\twocolumn[1][]{#1}%
\maketitle
\thispagestyle{empty}

\begin{center}

    \centering
    \includegraphics[width=\textwidth]{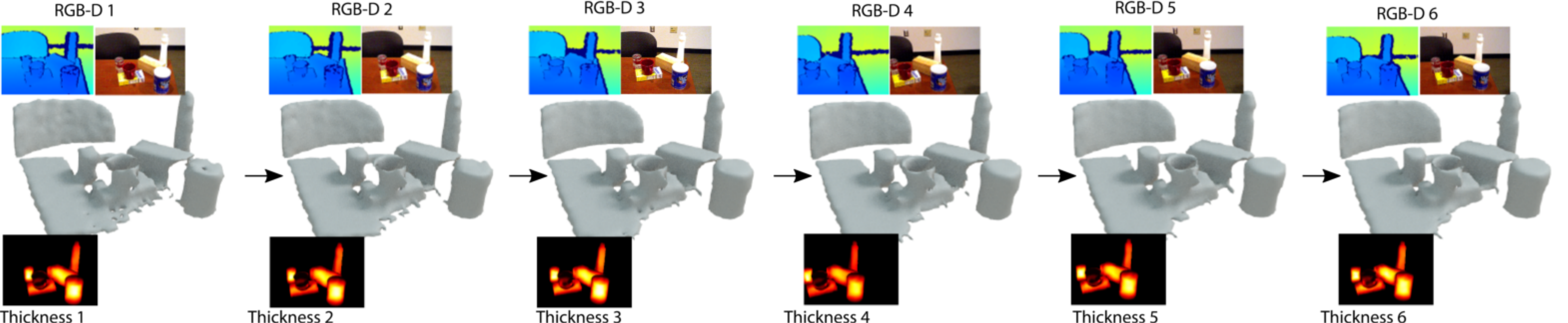}
    \captionof{figure}{Our approach uses predictions of the objects cross-sectional thickness to improve volumetric reconstruction quality. Top row shows the input to the proposed pipeline, an RGB-D frame. Bottom, cross-section prediction. From left to right in the middle, incremental reconstruction via our enhanced TSDF fusion algorithm. }

\end{center}%
}]

\begin{abstract}
    Detailed 3D reconstruction is an important challenge with application to robotics, augmented and virtual reality, which has seen impressive progress throughout the past years. Advancements were driven by the availability of depth cameras (RGB-D), as well as increased compute power, e.g.\ in the form of GPUs -- but also thanks to inclusion of machine learning in the process. Here, we propose X-Section, an RGB-D 3D reconstruction approach that leverages deep learning to make object-level predictions about thicknesses that can be readily integrated into a volumetric multi-view fusion process, where we propose an extension to the popular KinectFusion approach. In essence, our method allows to complete shapes in general indoor scenes behind what is sensed by the RGB-D camera, which may be crucial e.g.\ for robotic manipulation tasks or efficient scene exploration. Predicting object thicknesses rather than volumes allows us to work with comparably high spatial resolution without exploding memory and training data requirements on the employed Convolutional Neural Networks. In a series of qualitative and quantitative evaluations, we demonstrate how we accurately predict object thickness and reconstruct general 3D scenes containing multiple objects. 
\end{abstract}

\section{Introduction}
Knowledge of the shape of objects and of unseen part of the scene plays a critical role in applications such as robotic manipulation and autonomous exploration. In robot manipulation, the understanding of object geometry clearly influences the choice of grasping points. Similarly, in autonomous navigation, any additional information about occupied versus free space in the scene is helpful. The fusion of unseen information in the mapping process leads to more efficient exploration and faster map coverage.

Recent advancements in machine learning have fuelled improvements in single view 3D reconstruction. However, the developed techniques are not necessarily readily integrated with state of the art spatial mapping systems. 

In this work, we propose a novel approach to object reconstruction embedded in a scene that allows scalable multi-view reconstruction of both individual objects and groups thereof. The task we propose is to predict the geometry behind sensed surfaces in the form of view-centred cross-sectional thickness. We embed the thickness prediction network, X-Section, in a pipeline that allows to scale our approach to scene level. To integrate multiple views and recover 3D geometry, we suggest a modification to truncated signed distance function (TSDF) fusion.  Furthermore, our framework can be easily paired with other mapping approaches such as Bayesian probabilistic mapping \cite{loop2016}.

There are several reasons to prefer 2D predictions rather than trying to estimate the full 3D shape in one shot. One of the main advantages is that predicting an image instead of a voxel grid avoids the explosion in the number of weights of the network. Moreover, the use of a reconstruction algorithm to recover 3D geometry loosens the coupling between the reconstruction resolution and the network prediction. In an extensive study of different types of learning-based reconstruction approaches \cite{shin2018pixels}, the authors also found that view-centred pixel-wise predictions generalise better to unseen classes than object-centred voxel-based models.

\begin{figure}[h!]

    \centering
    \includegraphics[width=0.7\columnwidth]{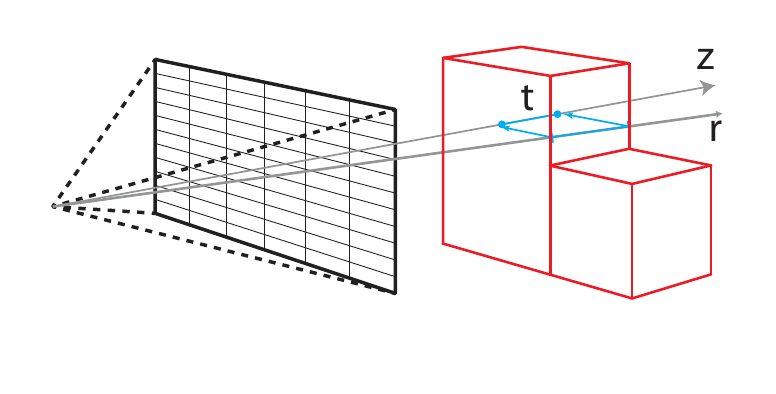}
    \vspace{-3mm}
    \caption{Illustration of the cross-sectional thickness. $t$ is the thickness of the surfaces hit by the ray $r$ and projected on the principal axis $Z$.}
    \label{fig:thickness_diagram}
\end{figure}

As obtaining training data for this task is challenging, we introduce a new dataset consisting of both synthetic and real images.  We render RGB, depth and thickness for models of the YCB Dataset \cite{calli2015} with domain randomisation. To achieve good performance on real data we fine-tuned on real sequences from \cite{xiang2017posecnn} with rendered thickness of the aligned objects. Similar to an X-Ray machine we render thickness by raytracing through synthetic models of objects and measuring the distance between the observed surface and the first surface behind it. An illustration of this cross-sectional thickness is shown in Figure \ref{fig:thickness_diagram}.

In short, we claim the contribution of our work to be fourfold:

\begin{itemize}
    \item A novel task to predict view-dependent 2D per-pixel thickness that can be used to efficiently recover a 3D volume.
    \item A complete pipeline from RGB-D or depth and silhouette (DS) to a full 3D reconstruction for 3D tabletop scenes using predicted thickness.
    \item A dataset of thickness data for 106k synthetic plus 34k real views of YCB objects, along with the RGB, depth and silhouette images and the code to render more views. 
    \item Training and prediction code with pre-trained weights to reproduce results.
\end{itemize}

The structure of the paper is as follows. We first review related works on  volumetric fusion, RGB-D shape completion and some single view RGB reconstruction approaches. We then introduce our approach and the dataset we train our model on. Finally we evaluate our model's performance on real RGB-D sequences.

\section{Related work}

\textbf{Surface Prediction and Spatial Mapping}
The most popular approach for reconstructing scenes from RGB-D images involves registering and fusing multiple frames into a 3D voxel grid. This volumetric fusion approach, popularised by KinectFusion \cite{newcombe2011}, works by first tracking the camera pose and then it uses the integration approach of Curless and Levoy \cite{curless1996volumetric} to fuse the depth images into the volume. Various improvements have been introduced, mainly focused on reducing tracking drift \cite{clark2018learning} and increasing the size of scenes that can be reconstructed. Kintinuous \cite{whelan2012kintinuous}, for example, uses a sliding volume to map large spaces. BundleFusion \cite{dai2017bundlefusion} reduces tracking drift by global bundle-adjustment and re-integration into the mapping process. \cite{vespa2018octree} tackles the efficiency bottleneck by means of a tree data structure. 
With the advent of deep learning there has been much interest in learning geometrical, structural and semantic priors to enhance the reconstruction process. For example, \cite{weerasekera2017dense} makes use of surface normal predictions to improve a monocular reconstruction. \cite{sunderhauf2017meaningful} uses semantic segmentation along with RGB-D reconstruction to create annotated maps of indoor scenes. More recently, Fusion++ \cite{mccormac2018fusion++} introduced an object-centric approach to large scale mapping which builds a map consisting of multiple TSDFs, each representing a single object instance.

\textbf{Volume Completion} A number of approaches propose to complete the scene starting form RGB-D information. Song et al.\ \cite{song2016ssc} and ScanComplete \cite{dai2018scancomplete} infer the missing voxels in a grid map along with the semantic labels. OctNetFusion \cite{riegler2017octnetfusion} describes a deep learnt fusion process using an octree data structure for efficiency. Their scheme can be seen as learning an implicit surface from the depth maps, helping with noise reduction and outlier suppression when fusing. Voxlets \cite{firman2016structured} operates on partially reconstructed 3D voxel grids. Other approaches \cite{yang20173d} use GANs to train an RGB-D to voxel predictor. The main disadvantage of these approaches is that it is inefficient for fusing multiple views as its 3D convolutions are both memory and compute intensive, restricting their use in real time applications.

\textbf{Silhouette based reconstruction} Shape-from-silhouette methods reconstruct the 3D shape of an object using multiple silhouette images taken from different viewpoints \cite{baumgart1975polyhedron}.

More closely related to our approach is \cite{prasad2005fast}, where the authors extract curves along the silhouette and reconstruct the object by finding the smooth surface which adheres to the edge curves. This method, however, requires that the object is symmetric and that the silhouette image is taken perpendicular to the symmetry axis.

\textbf{Single-view 3D reconstruction} Classical approaches to single-view reconstruction \cite{prasad2006single, criminisi2000single,hoiem2005automatic,horry1997tour,zhang2002single} relied on strong geometric priors. While these methods showed some impressive results on simple scenes, they lack ability to capture the complexity of real object shapes. 

The advent of deep learning has led to a major boost in the complexity and quality of scenes and objects that can be reconstructed from a single view. Approaches like  \cite{choy20163d,sharma2016vconv,tulsiani2018multi,kurenkov2017deformnet,girdhar2016learning,wu2016learning, gadelha20173d,zhang2018genre,brock2016generative,gwak2017weakly,tulsiani2018multi} all attempt to reconstruct 3D objects from 2D views and/or silhouettes. In the best case, these methods provide a view-centred reconstruction requiring to recover the translation and scale of the object, a challenging task itself. In the case the prediction is in a canonical pose, the full pose and scale has to be estimated. 

In a concurrent work, \cite{shin2019multi} represents an indoor scene as four layers of depth. Apart from the first, the layers of depth represent the full extension of an object along the ray. This might create artefacts in the case of non-convex shapes. Our work differs in the definition of the thickness as the distance between the observed surface and it's back and compensate for the incomplete representation of the geometry by means of integration with a multi-frame depth fusion algorithm.

\begin{figure}[h!]
    \centering
    \includegraphics[width=0.7\columnwidth]{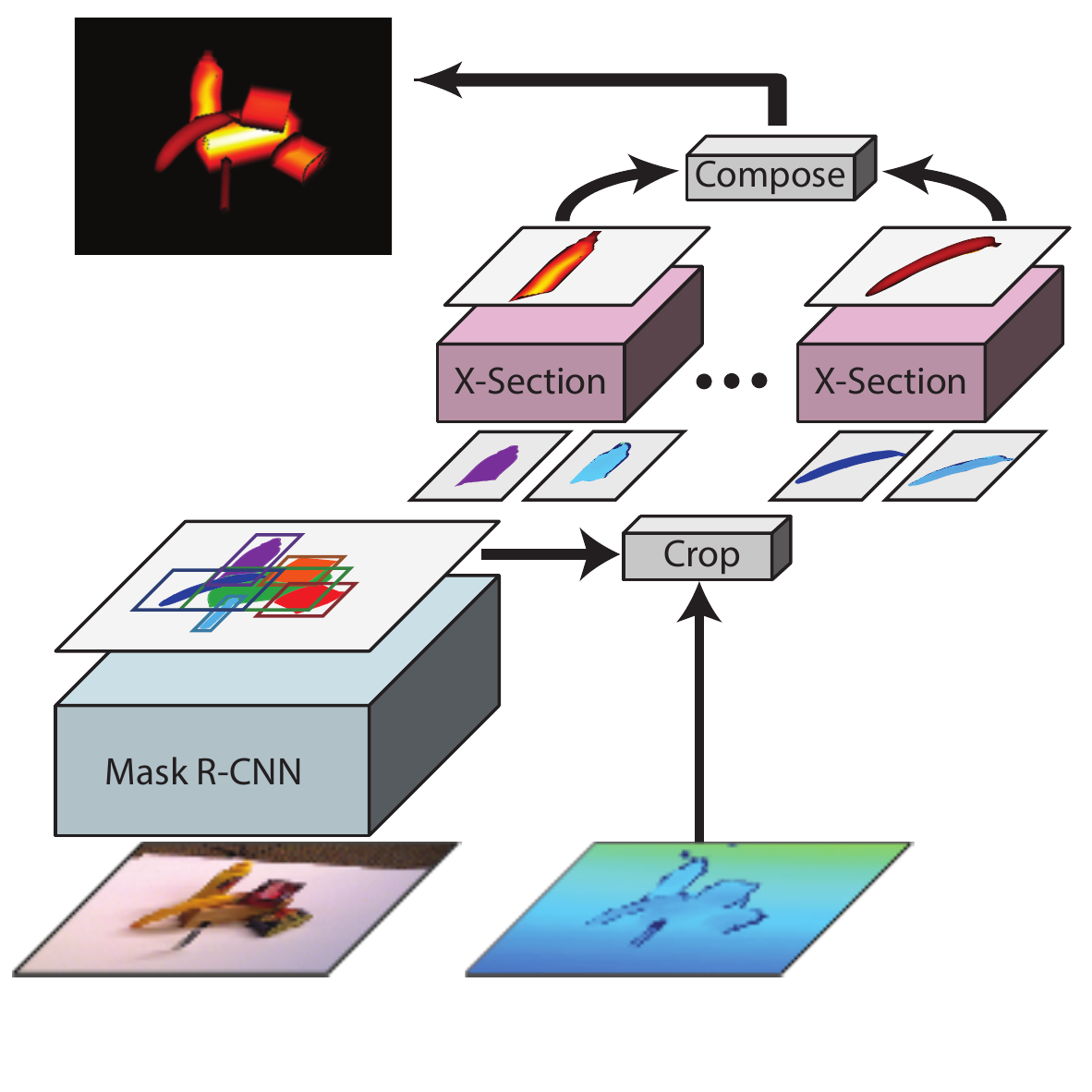}
    \caption{Overview of our cross-section prediction pipeline. An RGB frame is passed to Mask R-CNN. The resulting bounding boxes and masks are used to process RGB and depth data and crop single objects. X-Section is run for every object and the outputs are composed in a thickness frame.}
    \label{fig:architecture}
\end{figure}

\vspace{-0.5cm}
\section{Approach}
Predicting the thickness for an entire scene is a very demanding problem. Our method is based on the idea that decomposing this complex problem into smaller and simpler tasks makes the solution easier to find. We first decompose the scene into object instances and then produce an estimate for every object in the image. We then compose multiple predictions into a single frame that can be used in the fusion process to obtain a 3D model of the scene. 

As can be seen in Figure \ref{fig:architecture}, our system consists of five steps in total. An object detector, a pre-processing stage, a prediction operation and a final composition followed by a fusion step.

First, an object detector takes as input an RGB frame and outputs a set of bounding boxes and masks -- we use an off-the-shelf solution for this.  At the second stage of the pipeline, the output of the object detector is pre-processed to be input to our estimation network. The X-Section network is run for every object. Finally, the per-object predictions are merged in a single thickness frame and passed to the reconstruction algorithm that outputs a representation of the volume in a voxel grid.

\subsection{Object Detection and Instance Segmentation}
Our approach relies on any object detector that provides bounding boxes along with a segmentation masks of the object. For the current work, we chose an off-the-shelf version of Mask R-CNN \cite{he2017maskrcnn} based on ResNet \cite{he2016renset} and trained on the MS-COCO dataset \cite{mscoco}. Alternatives to Mask R-CNN include MaskLab \cite{chen2017masklab} or DCAN \cite{chen2017dcan}.

\subsection{Pre-processing}
The output of the object detector has to be pre-processed before moving to the estimation stage. We expand the bounding boxes to have a 4:3 shape ratio and use them to obtain RGB and depth patches along with corresponding silhouettes. To bridge the gap between the training and test depth images, we subtract the mean of the object region and the mean of the background to the corresponding pixels. In this way we aim to push the network to focus only on the shapes rather than on the absolute depth values. Images from a depth sensor are typically incomplete. At test time, we run an additional inpainting step, described in \cite{telea2004inpaint}, to recover missing data due to sensor noise.

\subsection{Thickness Network Architecture}
\begin{figure}[h!]
    \centering
\includegraphics[width=\columnwidth]{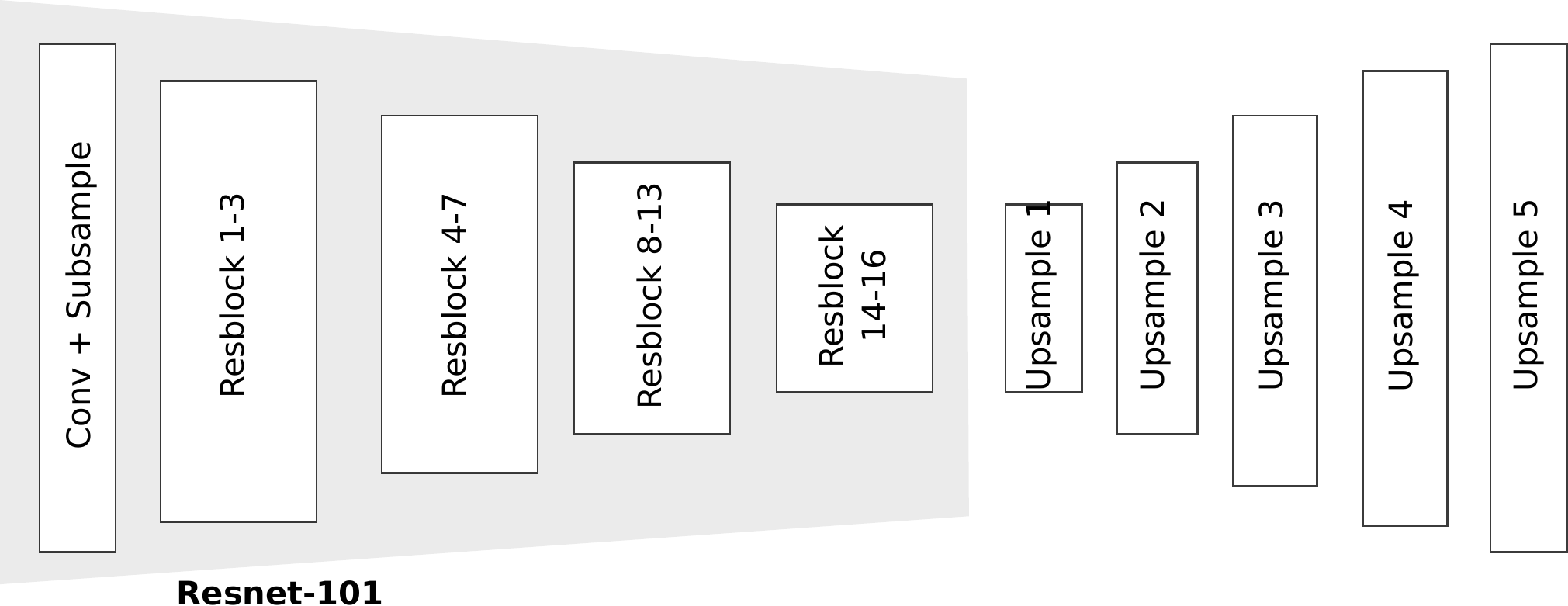}
\vspace{1mm}
    \caption{X-Section consists of a ResNet encoder and 5 upsampling blocks. The first layer blends the input in a 3-channel stack  used  by the encoder. Each upsampling block is composed from \textit{bilinear upsample - conv1 - conv2}. We use no skip layers apart from the residual connections in the encoder.}
    \label{fig:network}
\end{figure}

The network we propose to estimate thickness has an encoder-decoder structure in which input images are reduced to a code of dimension 3x4 with 2048 channels. Considering the affinity of our task with object recognition and given the limited size of the available dataset, we use an encoder based on ResNet with pre-trained weights on ImageNet. Since our input differs from the original one the network was trained with, we add an additional convolutional layer that takes stacked depth and silhouette images (or RGB and depth) and outputs a 3 channel feature image. The decoder consists of blocks of upsampling followed by two convolutional layers with ReLu \cite{nair2010relu} activation along all the layers except for the last one, which is linear. There are no skip connections between the encoder and the decoder part of the network. We train by minimising the $\mathcal{L}_2$ loss between the predicted and the ground truth thickness. Figure \ref{fig:network} depicts an example architecture based on ResNet101.

\vspace{-0.3cm}
\subsection{Enhanced TSDF Fusion} \label{sec:tsdf}
2D thickness prediction can be used to recover the 3D shape by fusing multiple frames, or even form a single view. To do so, we introduce an enhanced 3D fusion algorithm based on the approach of Curless and Levoy \cite{curless1996volumetric}. The affinity of the thickness signal to depth measurement allows for easy integration into existing frameworks. 
\vspace{-0.2cm}
\begin{figure}[h!]
    \centering
    \includegraphics[width=\columnwidth]{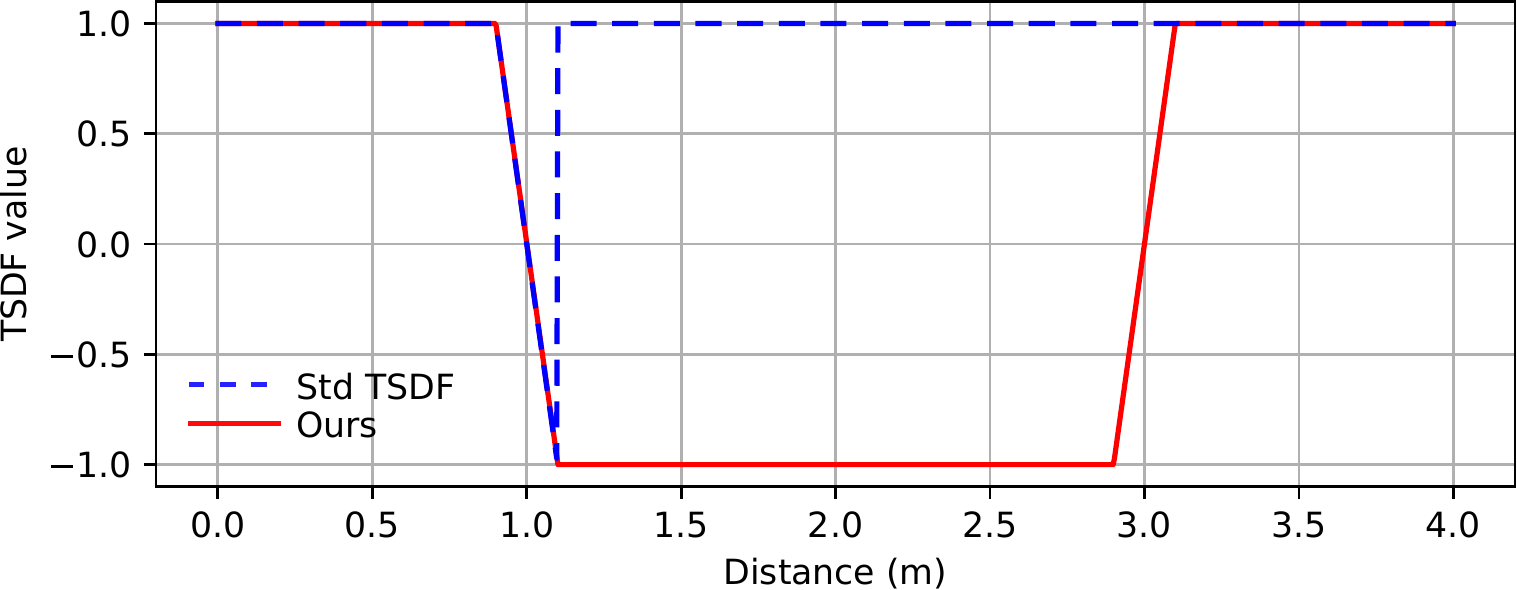}

    \caption{A plot of our thickness enhanced TSDF and standard TSDF. We show an example of surface at 1.0m with thickness 2.0m.}
    \label{fig:TSDF_thickness}
\end{figure}


The value of the new TSDF $\phi(z)$ depends on the truncation value $\tau$ that define the margins in which front and back surfaces lie respectively; $d$ and $t$ denote the depth and thickness value at a pixel $\mathbf{u}$ and $z$ the position along the ray of the camera corresponding to that pixel : 
\begin{equation}
    \phi(z) = 
    \begin{cases}
    1                       &\quad  z \leq d - \tau,                     \\
    \frac{d - z}{\tau}      &\quad  d - \tau < z < d + \tau,             \\
    -1                      &\quad  d + \tau \leq z \leq d + t - \tau,   \\
    \frac{d + t - z}{-\tau} &\quad  d  + t - \tau < z < d + t + \tau,      \\
    1                       &\quad  z \geq d + t + \tau.  
    \end{cases} 
\end{equation}

The resulting TSDF profile is shown in Figure \ref{fig:TSDF_thickness}. In contrast to methods such as \cite{newcombe2011} this reconstruction algorithm does not only yield surfaces, but explicitly reconstruct the occupied volume of an object. Multiple frames are fused by weighted average of the TSDF for each frame. When a voxel is updated the corresponding weight is incremented.

\section{Dataset} \label{sec:dataset}
In order to generate thickness data we need a dataset with a complete model of each object. Most large-scale RGB-D datasets \cite{silberman2012,song2016ssc,InteriorNet18} provide 2D images with depth and object instances but do not provide full 3D data about the objects. A dataset that  satisfies this requirement is the YCB dataset \cite{calli2015}. YCB is composed of 92 objects belonging to 77 classes. The dataset provides water tight meshes with textures extracted from images. 

\begin{figure}[h!]
    \centering
    \includegraphics[width=\columnwidth]{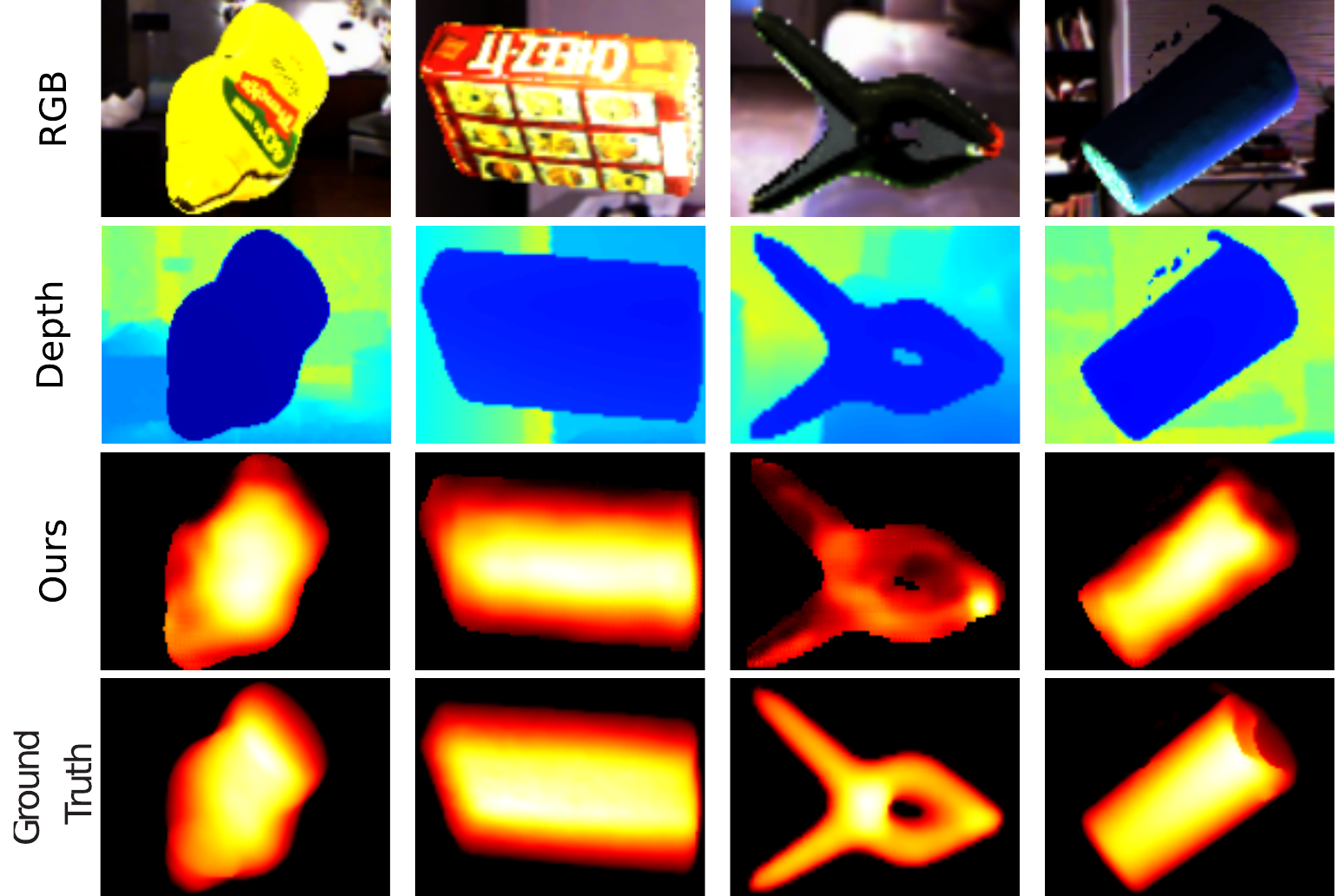}
    \caption{Examples of training data with prediction of the synthetic YCB dataset. Objects are hard to recognise because of domain randomisation and subsampling. Thickness is predicted by one of the fully trained networks of which the performances are reported below. }
    \label{fig:synth_training_data}
\end{figure}

\cite{sundermeyer2018implicit} suggests that randomisation of certain attributes leads to the robustification of the learning with respect to that characteristic. Hence, we render objects with random number of lights, intensity, colour and positions. This domain-randomisation approach aims to guide the network to ignore environmental features and focus on shape cues. 

\begin{figure*}[h!]
    \centering
    \includegraphics[width=\textwidth]{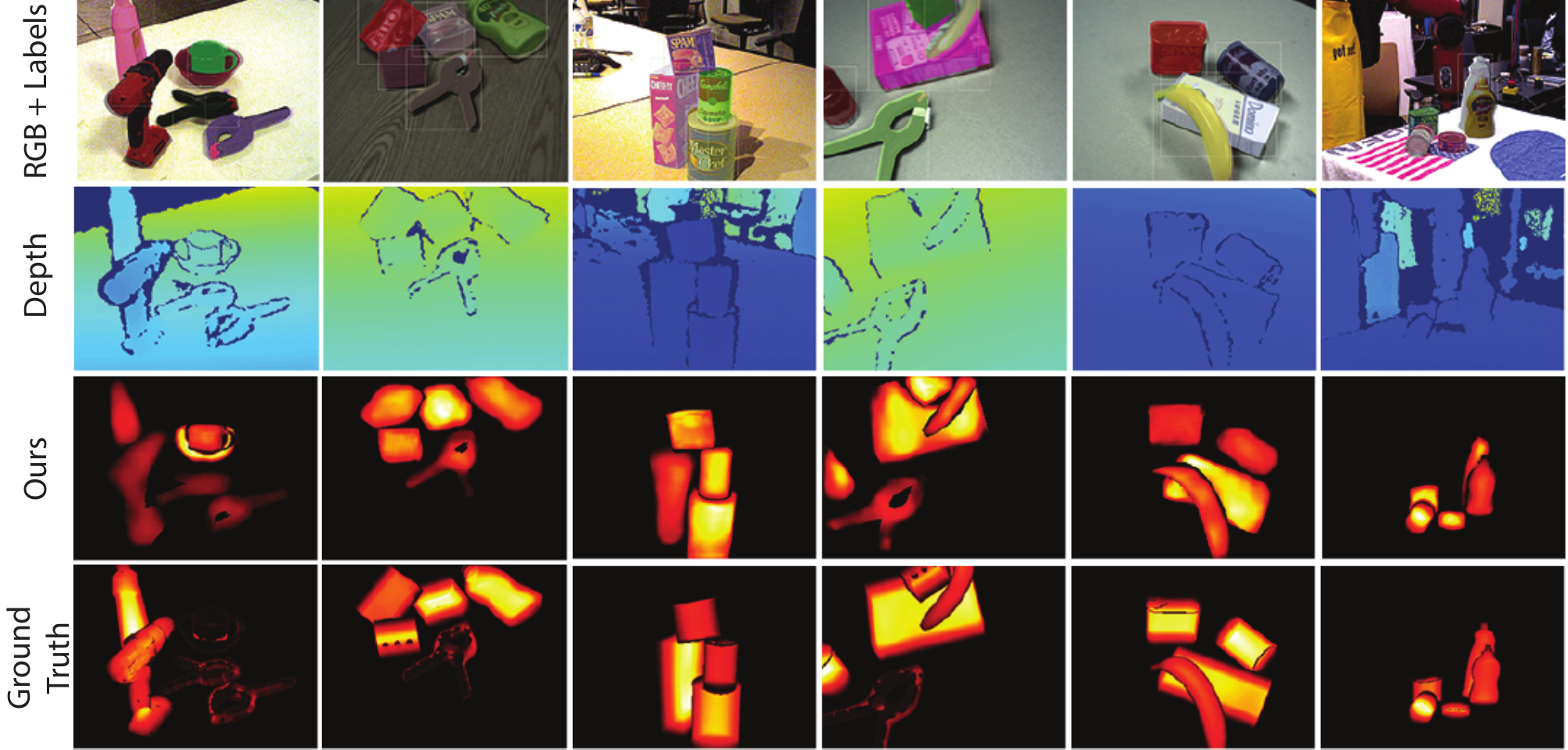}

    \caption{Prediction on the YCB Video dataset using ground truth bounding boxes and segmentation.}
            \vspace{-3mm}

    \label{fig:ycb_video_predictions}
\end{figure*}
Our rendering pipeline renders depth and RGB at a resolution of $640 \times 480$ with objects at a random distance from the camera. We then crop the image using the bounding box of the object and resize the crop using bilinear sampling, simulating the object detection process. To add  more realistic background we placed the rendered object in front of RGB and depth frames randomly picked from the NYU dataset \cite{silberman2012}. The resulting dataset comprises 2000 images per modality for 86 of the objects in the YCB dataset. Figure \ref{fig:synth_training_data} shows a sample of the training dataset along with network prediction and ground truth cross-section. 
\vspace{-0.1cm}
Cross-sectional thickness is rendered with a custom shader in Blender\footnote{https://www.blender.org/}. By design the shader returns only the visible surface thickness. Subsequent surfaces are ignored. This choice is inspired by our focus on multi-view fusion. Our approach allows for the incremental refinement of an object by fusing predicted thickness over multiple views. By not predicting the thickness of unobserved surfaces, we avoid integrating wrong information from hallucinated structures.

To bridge the gap between real and synthetic data, we fine tune the network on the YCB Video dataset presented in \cite{xiang2017posecnn}. The dataset is composed of 90 videos of table top scenes captured with an Asus Xtion Pro Live. Every RGB and depth image is accompanied by semantic labels, bounding boxes and poses of the objects relative to the cameras. We take advantage of such information to replicate the scene in Blender and render the thickness frame. We then use bounding boxes and labels to crop patches of single objects from depth and thickness and to create the corresponding silhouettes. In this way we render 100 thickness images for each of 80 of the videos.

\section{Results}
To analyse the effectiveness of the approach, we trained X-Section and design three experiments. In 2D we compare against the validation set. Since our method predicts unseen information form RGB-D frames, it can be seen as a shape completion problem. Hence, we benchmark our pipeline against Voxlets \cite{firman2016structured}. Finally, we fuse multiple predictions and show the difference with respect to a voxelised representation of the scene. 

The ResNet backbone is pre-trained on ImageNet and the whole network is trained for 40 epochs, with learning rate of $1e-5$ and batch of 50, 128x92 images. We reserve ten percent of the dataset as validation set. The model is then fine-tuned on data from YCB Video leaving out 12 sequence for validation. We found 10 epochs to be sufficient to achieve satisfactory results.

\begin{table*}[h!] 
\centering
\vspace{1mm}
\begin{tabular}{|l|c|c|c|c|c|}
\hline
                               &          &\multicolumn{4}{c|}{Ours} \\
\hline
                               &          &\multicolumn{2}{c|}{ResNet 101} & \multicolumn{2}{c|}{ResNet 50} \\
\hline
                               & Baseline & DS     & RGB-D   &  DS             & RGB-D    \\ \hline
\hline
absolute relative difference   & 96.044   & \textbf{3.819}  & 4.301  & 3.896  & 4.047 \\ 
sqr relative difference        &  4.074   & 0.047  & 0.056  & \textbf{0.045}  & 0.059 \\
RMSE (linear)                  &  0.026   & 0.015  & 0.015  & \textbf{0.013}  & 0.014 \\
RMSE (log)                     &  1.545   & 0.700  & 0.693  & \textbf{0.671}  & 0.689 \\
\hline
\end{tabular}
\caption{2D evaluation results on the YCB-video dataset. Thickness is measured in meters. We test different inputs, Depth with Silhouette (DS) and RGB with Depth (RGB-D). The baseline is the mean thickness over the training dataset.}
\label{res:2d}
\end{table*}

\subsection{2D Evaluation}
To the best of our knowledge there is no related method that has been proposed to predict the cross-sectional thickness of objects. Thus, we adopt the mean thickness over all pixels of the objects in the training set as reference. We test two variants of X-Section, one with ResNet50 and with ResNet101 backbone. Both networks are trained on the same amount of data for the same number of epochs. We define $t_p$ and $\hat{t}_p$ as the ground truth and predicted thickness, respectively. Over $N$ pixels we compute the metrics:

\begin{align} \label{math:metrics}
    \text{Abs. Relative Difference} &= \frac{1}{N} \sum_{p} \frac{| t_p - \hat{t}_p |}{t_p},    \\ 
    \text{Square Relative Difference} &= \frac{1}{N} \sum_{p} \frac{\|t_p - \hat{t}_p \|^2}{t_p},    \nonumber  \\
    \text{Log Root Mean Square} &= \sqrt{\frac{1}{N} \sum_{p} \| \log{t_p} - \log{\hat{t}_p} \|^2}.  \nonumber   
\end{align}

The results are gathered in Table \ref{res:2d}. As expected the network performs better than mean on all tests. It can be noticed that the performance gap between two different versions of X-Section is not significant. This hints that breaking down the scene in smaller components simplifies the task, requiring a smaller network. A more thorough investigation is required to draw conclusions about this and left as future work.
Large values of the \textit{absolute relative difference} of the baseline are the result the view-centred formulation of the task that makes the data dependent on the incidence angle of the observation ray. As a consequence the value of thickness tends to zero at the border of objects where rays are tangent to the surface. The fact that X-Section produces such low values for this metric suggests that the network has actually learnt to predict the shape coherently.

\subsection{RGB-D Vs. Depth and Silhouette} \label{sec:ablation}
To isolate where most of the information is stored, we have trained a network with RGB and depth as input and one with a depth image and a silhouette. As shown in Table \ref{tab:3D} and Table \ref{res:oneseq} the use of RGB and depth causes a drop in performance. When a mask is passed in input, the network takes the mask into account when making predictions and this guides the learning to better exploit the information stored in the pixels picturing the object. 

Although in principle the RGB data should hold important information for shape reconstruction, this type of input is the one that suffer from domain adaptation the most. It is also to be considered that depth retains direct information of the shape and it might cause the network to ignore cues in colour data. This analysis leans in favour of the use of 2.5D sketches for shape recovering. However, a stronger conclusion on the best input for this type of algorithms requires a more thorough and precise analysis that is out of the scope of this work. 

\subsection{Comparison with Voxlets} \label{sec:3deval}

Our focus is to retrieve geometric information from an incomplete measurement of the environment. This makes this work closely related to 3D shape completion, such as \cite{dai2018scancomplete} or \cite{firman2016structured}. The voxel resolution of the former approach is 5cm making it hard to directly test it in table top scenarios. On the contrary, Voxlets \cite{firman2016structured} is showcased in table top scenes and provides trained models and data.

\begin{table*}[h!] 
\centering

\vspace{1mm}
\begin{tabular}{|l|c|c|c|c|c|c|}
\hline
 & \multicolumn{2}{|c|}{Baselines} & \multicolumn{4}{|c|}{Ours (X-Section)} \\
\hline
          &         &              &\multicolumn{2}{c|}{ResNet 101 base} & \multicolumn{2}{c|}{ResNet 50 base} \\
\hline
          & Voxlets & DF & DS     & RGB-D  &  DS    & RGB-D \\
\hline\hline
IoU       &  0.713  &  0.327       & \textbf{0.761 } & 0.620 & 0.759  &  0.651 \\
Precision &  0.893  &  0.887       & \textbf{0.894}  & 0.875 & 0.837  &  0.882 \\
Recall    &  0.779  &  0.341       & \textbf{0.836}  & 0.680 & 0.890  &  0.713 \\
\hline
\end{tabular}
\caption{Results of the comparison against Voxlets \cite{firman2016structured} for sequences with all objects detected. As baseline we adopt Voxlets and our implementation of a depth only fusion algorithm via TSDF averaging (DF).}
\label{res:oneseq}
\end{table*}

We run our pipeline on the dataset released with \cite{firman2016structured} and we pick eight scenes with highest detection rate. As ground truth we use the voxel grids provided. Most of the instances are completely new to the network and their shape non trivial. Examples of objects of the dataset are boxes, shoes, a teapot and a cast head.  We think this difficult scenario thoroughly tests the generalisation capabilities of the network.  We run our pipeline on a single frame and compare our single-view reconstruction with the 3D completion approach in Voxlets. Figure \ref{fig:voxlet_comparison} shows the scene reconstructed with our method, our implementation of a depth only fusion algorithm, the output of Voxlets and reference complete volume. 

\begin{figure*}[t!]
    \centering
    \includegraphics[width=\textwidth]{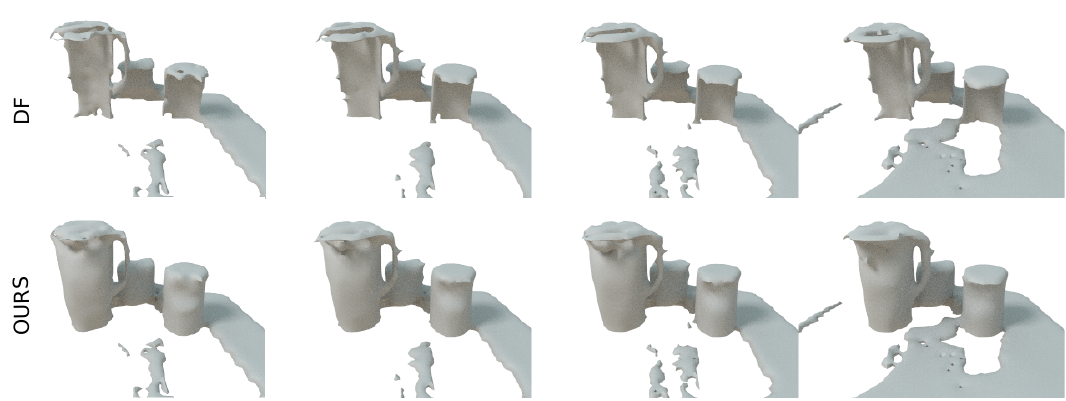}
    \caption{Proposed enhanced fusion in a YCB Video sequence. Top row, fusion of depth frame with TSDF averaging (DF). Bottom row, the proposed augmented fusion. We chose spatially distant frames. From left to right, fusion of frame 0, 60, 120 and 270.}
    \label{fig:multiview}
\end{figure*}
After fusing the predictions in a TSDF volume using the algorithm described in Section \ref{sec:tsdf}, we recover occupancy values by binarising the obtained TSDF values in the 3D grid. We classify voxels as occupied if the TSDF values are less than the truncation value $\tau$ and free otherwise. Calling $\mathcal{V}_g$ the ground truth volume and $\mathcal{V}_x$ the volume reconstructed with X-Section predictions, Intersection Over Union, precision and recall can be computed as

\begin{equation}
    IoU = \frac{\mathcal{V}_g \bigcap \mathcal{V}_x}{\mathcal{V}_g \bigcup \mathcal{V}_x}, \quad
    P = \frac{p_t}{p_t + n_t} \nonumber , \quad
    R = \frac{p_t}{p_t + n_f}. \nonumber
\end{equation}
Where $p_t$ is the number of true positive predictions (so a voxel correctly predicted as belonging to the object volume), $n_t$ denotes the number of true negatives and $n_f$ the number of false negatives. 

Table \ref{tab:3D} shows X-Section falling short of few percentage points with respect to the baseline. There are several reasons behind the accuracy of our approach on this data. A crucial factor is that the objects used for this benchmark do not compare to the ones in the dataset, hence the network is seeing not only a novel view, but also a novel model and novel class for all inputs. Moreover, our approach does not complete the scene where there are no depth readings. This yields to incomplete reconstruction when objects are occluded. On the other hand, Voxlets tries to fill the gaps, scoring better in the chosen metrics.

To investigate the impact of a faulty object detector, we ran the pipeline on a sequence where all objects are successfully segmented. As Table \ref{res:oneseq} shows, in this case the accuracy of the prediction is beyond what Voxlets achieves; showing impressive generalisation capabilities. The use of an object detection stage results in a trade off in terms of generality. Isolating the single objects is portable across different scenarios and environments without requiring any retraining or fine tuning. Voxlets, however, needs to be trained on every different scene type.

\subsection{Multi-Frame Fusion Evaluation}
The main application of the X-Section pipeline is the integration of thickness prediction in a multi-frame fusion system. The  YCB Video dataset \cite{xiang2017posecnn} provides relative poses of object with respect to the camera. We use this information to compose the scene and produce a solid voxelisation to be used as ground truth approximation. Figure \ref{fig:ycb_video_predictions} shows the result of our pipeline on sample frames of the validation dataset. Using the algorithm in Section \ref{sec:tsdf} we fuse the predictions for the first 50 frames of each of the 12 validation sequences.

\begin{table*}[h!]
\centering

\begin{tabular}{|c|c|c|c|c|c|c|c|c|c|c|c|c|}
\hline
           &  \mc{2}{0048} & \mc{2}{0049} & \mc{2}{0050} & \mc{2}{0051} & \mc{2}{0052} & \mc{2}{0053}\\ 
\hline
           & DF  & Ours & DF & Ours & DF & Ours & DF & Ours & DF & Ours & DF & Ours \\

\hline\hline
IoU        & 0.299 & \textbf{0.535} & 0.346 & \textbf{0.513} & 0.233 & \textbf{0.392} & 0.355 & \textbf{0.735} & 0.264 & \textbf{0.693} & 0.252 & \textbf{0.395} \\ 
Precision  & 0.787 & \textbf{0.841} & \textbf{ 0.745} & 0.659 &\textbf{ 0.872} & 0.804 & 0.894 &\textbf{ 0.901 }& \textbf{0.911}  & 0.881 & 0.484 & \textbf{0.535} \\ 
Recall     & 0.326 & \textbf{0.596} & 0.393 &\textbf{ 0.698} & 0.241& \textbf{0.433} & 0.371 &\textbf{ 0.780} & 0.271 & \textbf{0.764} & 0.345 &\textbf{ 0.600} \\ 
\hline
\end{tabular}
\caption{Evaluation of multi-frame fusion averaged over the first 50 frames of the YCB Video dataset \cite{xiang2017posecnn}. We compare our modified TSDF fusion of Section \ref{sec:tsdf} and a depth only fusion algorithm, labelled DF.}
\label{res:multiframe}
\end{table*}
\begin{figure}[h!]
\includegraphics[width=\columnwidth]{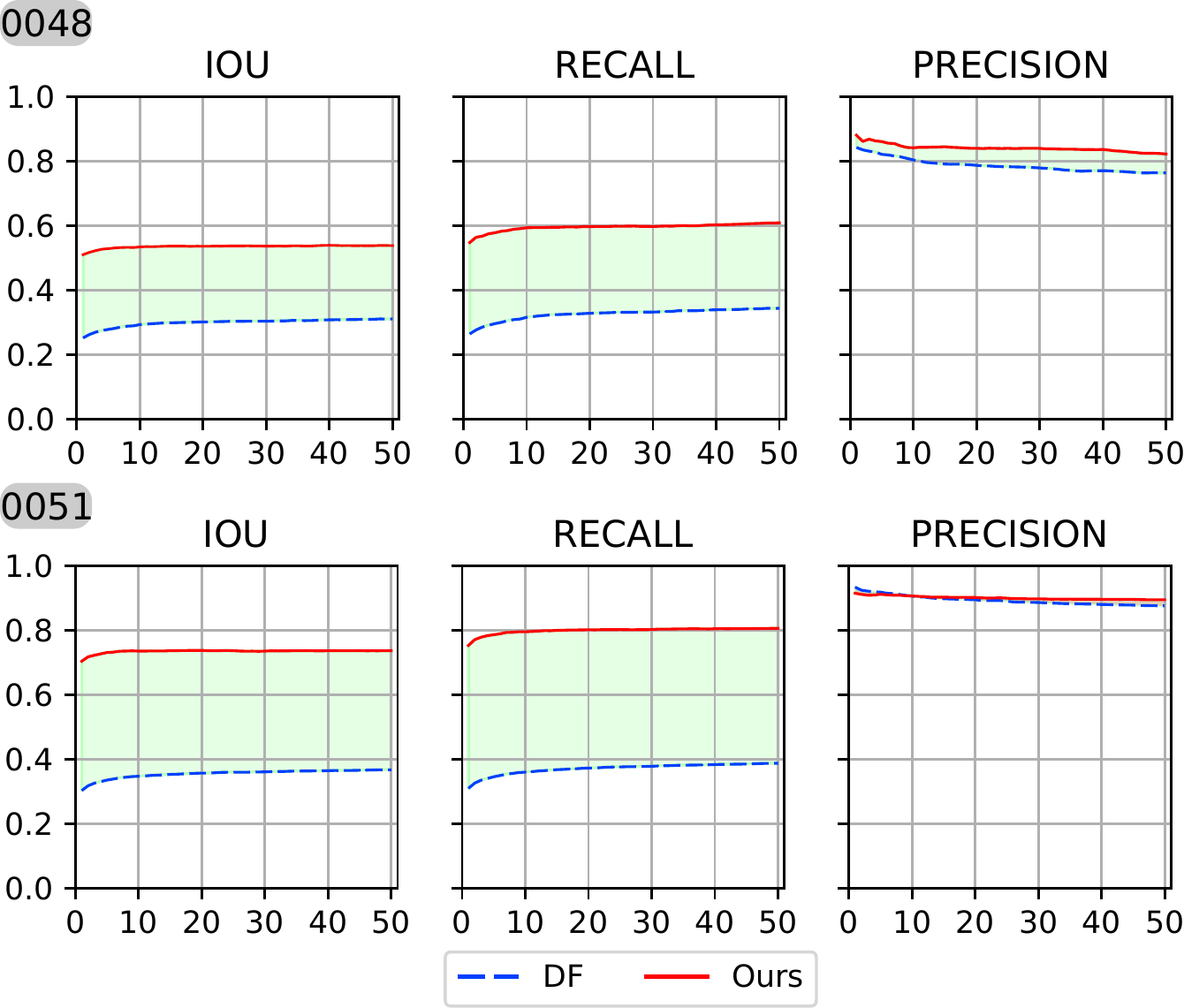}
\caption{Evaluation of multi-frame fusion for two of the sequences of the YCB Video dataset \cite{xiang2017posecnn} reserved for validation. Top sequence \textit{0048}, bottom \textit{0051}. The red solid line represent the result of our method, the blue dashed line shows the performance of a depth only fusion algorithm.}
\label{plots:multi_frame}
\end{figure}
Figure \ref{plots:multi_frame} reports the metrics computed per each frame fused from sequence \textit{0052} and \textit{0048} of the dataset. In this two scenes we report IoU and recall almost twice as high as the ones obtained by fusing only depth frames. This is a consequence of reconstructing explicitly the volume and not only the surface as traditional TSDF fusion algorithms do. However, it is also important that we recover accurately the shape of the object. This is reflected by the precision metric. On this specific case $90\%$ of the voxels recovered are true positives, matching the performance of depth only fusion that uses only sensor readings. 

Table \ref{res:multiframe} reports the average value for all the metrics for every validation sequence. IoU and Recall rates are always in favour of the suggested pipeline. On some sequences, our approach falls slightly short in terms of precision. Since the proposed method predicts unseen surfaces in difficult scenes the network predicts a small percentage of false positives. This drawback could be mitigated by predicting a per pixel uncertainty and use it for probabilistic mapping. Investigations in this direction are reserved for future work. 

Figure \ref{fig:multiview} reports the result of multiple-view fusions of another validation sequence. The reconstructed scene is shown from the back of the observed surfaces. The frames are relatively spatially distant for a table top scene. The bottom row shows the result of the thickness fusion algorithm described in Section \ref{sec:tsdf}. The results shows consistent predictions and over time the reconstruction quality improves. Whenever there is no thickness information (such has the table surface) only depth is fused (i.e.\ with traditional TSDF).

\begin{figure}[h!]
    \centering
    \includegraphics[width=\columnwidth]{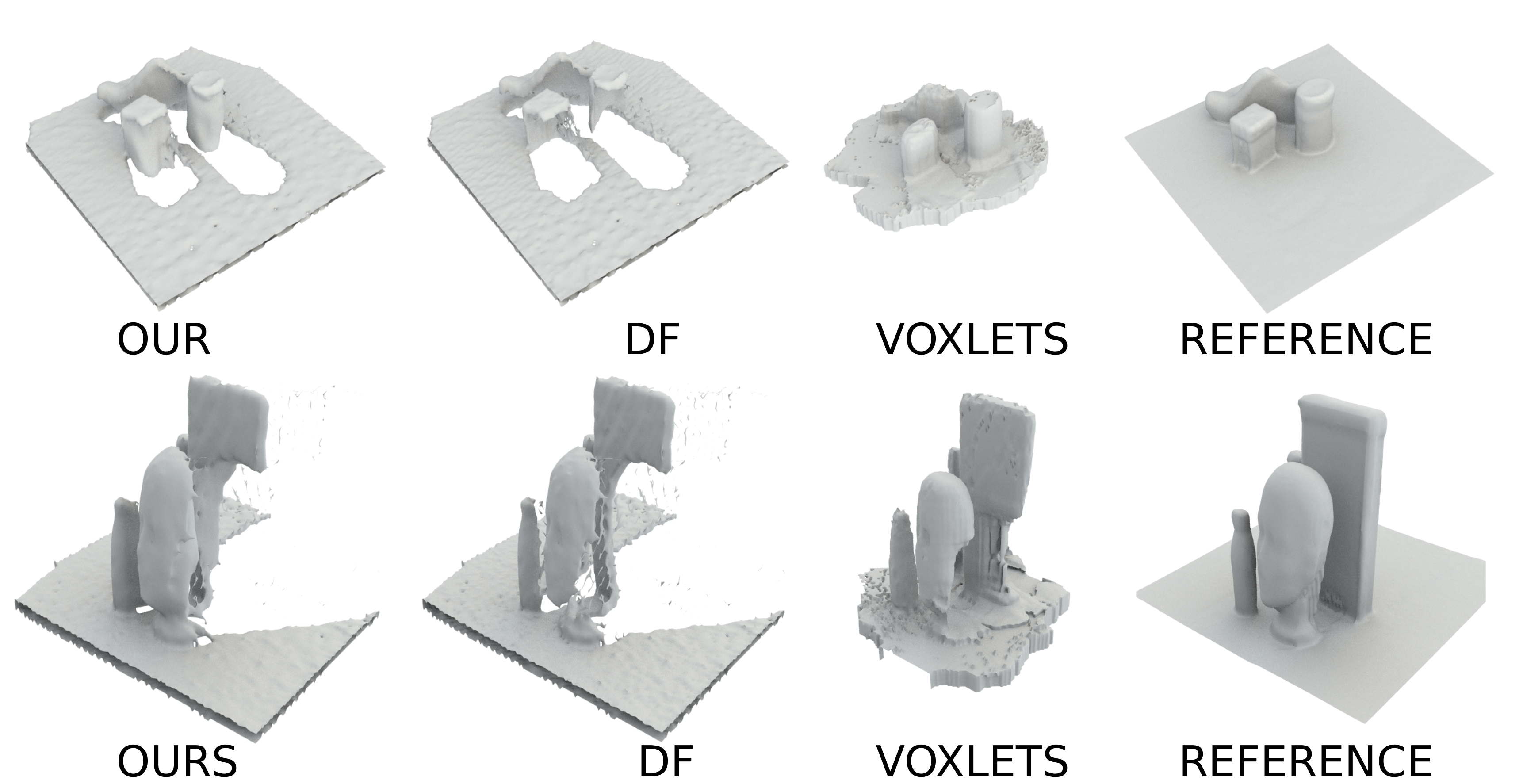}
    \caption{Reconstruction results and comparison with Voxlets. Each row shows two different reconstructed scene. From left to right: results of our fusion algorithm using predicted thickness, results of a depth only fusion via TSDF averaging (DF), the output of Voxlets and the reference model. }
    \label{fig:voxlet_comparison}
\end{figure}

\begin{table}[h]
\vspace{-2mm}
\centering

\vspace{1mm}
\begin{tabular}{|l|c|c|c|c|c|}
\hline
          & Voxlets &  DF  &{Ours }  \\
          &  &  & (ResNet 101 - DS)       \\
\hline\hline 
IoU       &  \textbf{0.622 } &  0.234       &  0.440  \\
Precision &  \textbf{0.811}  &  0.695       &  0.703   \\
Recall    &  \textbf{0.735}  &  0.261       &  0.536   \\
\hline

\end{tabular}
\caption{3D evaluation of our approach on the Voxlets dataset for eight sequences on which Mask R-CNN has missing detections. We show comparisons against Voxlets and depth only fusion via TSDF averaging (DF). }
\label{tab:3D}
\end{table}

\section{Conclusions And Future Work}
In this work we have presented the novel task of predicting the cross-sectional thickness of objects in a scene. We introduced a model for solving this task that involves decomposing a scene into individual objects, predicting the thickness and then recomposing the scene. Our experiments show that we can train our model and recover the 3D shape of the object with a simple extension to traditional fusion algorithms. To overcome the difficulties of domain adaptation we fine tuned on real world images. This proved to be central for test time performances.  

We demonstrated the convenience and compactness of predicting the cross-sectional thickness of objects and it's usefulness in reconstruction scenarios. Moreover, predicting one layer only has the advantage of limiting the estimation to observed surfaces, avoiding inaccuracy caused by the network hallucinating non observable parts of the scene. On the other hand this might yield incomplete models. There are different ways to approach this issue and we aim to investigate some in future works.

{\small

}



\section*{Supplementary material}

# X-Section: Cross-Section Prediction for Enhanced RGB-D Fusion. Supplementary Materials.

Andrea Nicastro[1], Ronald Clark[1], Stefan Leutenegger[2]
[1]Dyson Robotics Lab, [2]Smart Robotics Lab, Imperial College London
a.nicastro15@imperial.ac.uk

## Multi-Frame Fusion Results

This section expands further the multi-frame fusion experimental analysis. We obtain a ground truth approximation by computing a solid voxelisation of the scene. Cad models and objects poses are given by the dataset. Since it is not possible to obtain the ground truth of the environment (such as tables, walls and monitors), we mask the depth using segmentation information to reconstruct objects only.

We run the pipeline for the first 50 frames and fuse them. The resulting reconstruction is compared with the ground truth. Finally, we compute Intersection over Union, precision and recall for every frame and report them in Figure 2. The reconstructions are shown in Figure 3 and 4.

Overall, the results show that in all scenes we recover more information than depth only fusion. For most sequences the precision of the reconstruction using thickness is close to the one obtained by fusion of only depth. This means that our propose method recovers correctly the information and does not hallucinate details that are not present. However, there are two in which X-Section has a significantly lower precision than depth only fusion. Sequences *0049* and *0057* result particularly difficult to reconstruct.

As shown in the second row of Figure 3, among the objects in *0049* there is an upside-down bowl. In this case the network struggles to generate a consistent prediction outputting a different internal shape as expected. An detail of the reconstruction of the scene is reported in Figure 1. The same object is present also in sequence *0053* in which our method has really high precision and does not present any artefact. Since the network has been trained using random position of cameras and objects, it should be robust to changes in the point of view. Among the possible causes of this failure there is the occlusion between objects and the lack of context at training time.

The analysis above highlights one of the current drawbacks of the proposed approach. Isolating objects simplifies the task and yields effective generalisation to unseen views. However, this introduces a compromise in terms of robustness due to the lack of context. The impact remains limited, and tackling this issue is left as possible improvement.

Another problematic sequence is *0057*. The sources of error are twofold. One is a clear over estimation of the thickness of the objects (also present in *0049* as Figure 1 shows). The other is the presence of artefacts around the shape of the objects. These recur often in the dataset and they are typical of dense reconstruction systems. In the case of the proposed pipeline they are caused by drift in the pose and imprecise segmentation at the edges of object. The dataset used for evaluation has camera poses associated to each frame, but they are the result of global optimisation. This causes a pose drift that affects all the scenes. Both segmentation accuracy and pose drift are currently open research questions and the integration of more advanced methods addressing this issues will improve the performances of our method.

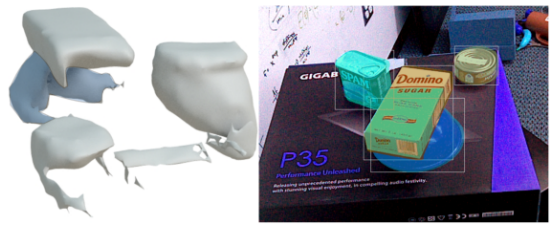

Figure 1: Reconstruction of Sequence *0049* using predicted thickness. Highlighted in blue the upside down bowl that causes the network to fail. On the right the corresponding object is segmented in the same colour.

In summary, the experiments show a measurable improvements in the estimation of the occupied space. The results also show that breaking down the problem produces good generalisation to both novel views and novel objects (see main body of the work for this) at the cost of loosing contextual information and of a compromise in terms of stability. However, we measured a marginal impact of the drawbacks on the validation sequences.

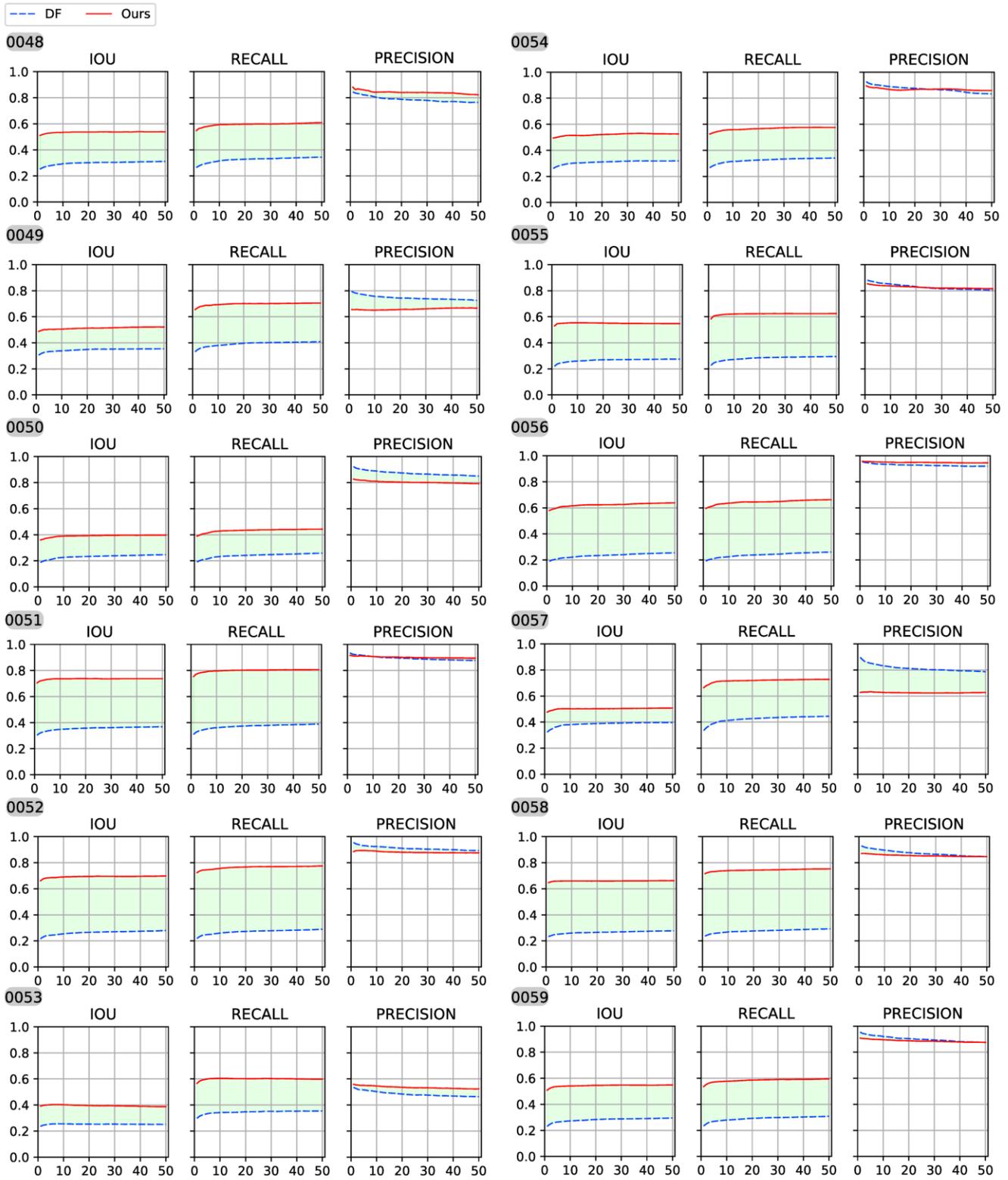

Figure 2: Plots of the metrics for each one of the validation sequence. The plots present tmetrics evaluated after the fusion of a single frame for the first 50 frames of the sequences. The solid red line is our approach while the blue dashed plot is obtained by fusing only depth information via TSDF averaging (DF).

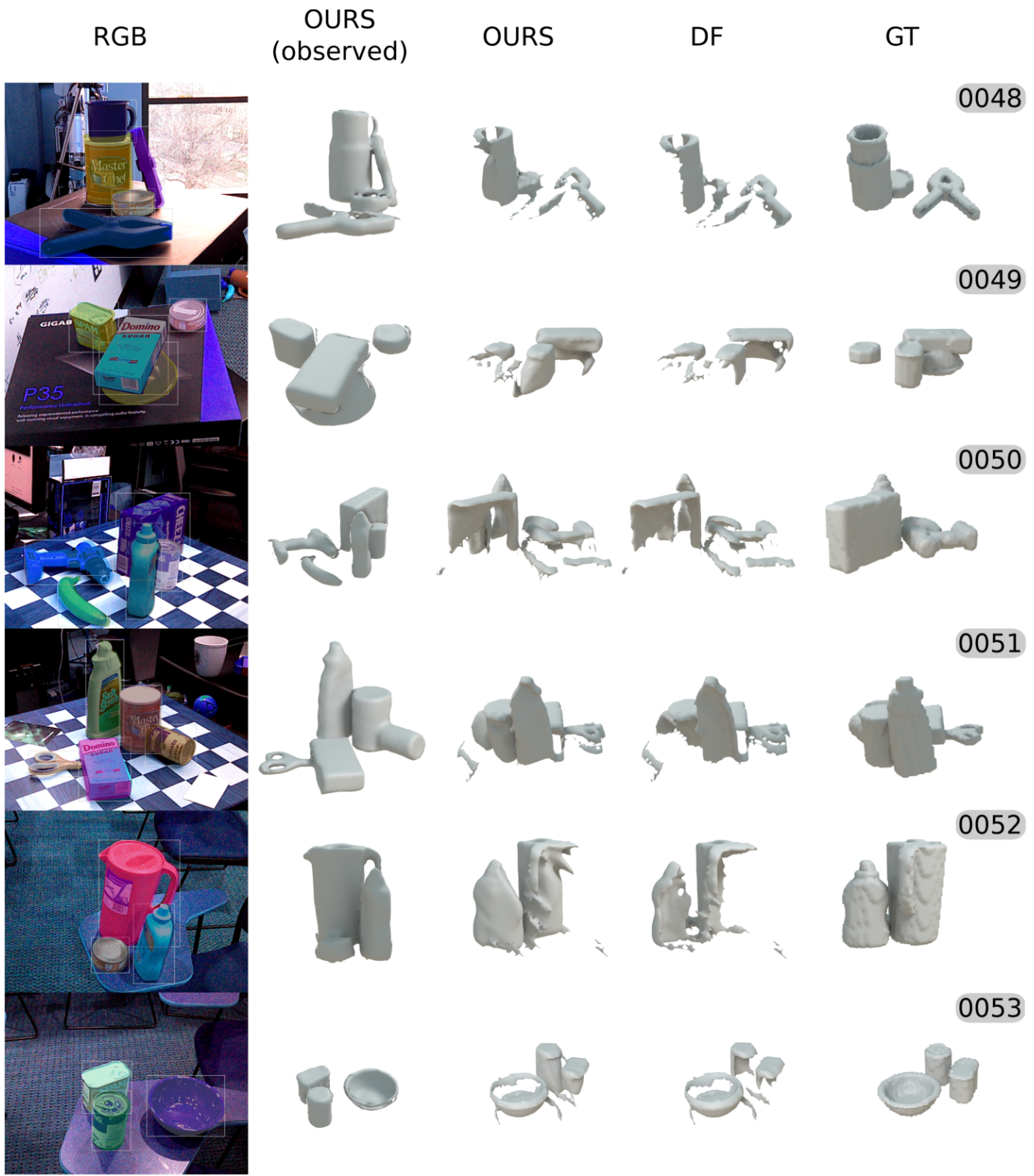

Figure 3: Reconstruction resulting from the fusion of the first 50 frames of the sequences. In this image are reported the first six sequences, *0048* to *0053*. From left, first RGB frame of the sequences, our reconstruction using thickness first showing the observed surfaces and then the estimated occupied space. The reconstruction labelled as DF is obtained by fusing only depth frames using traditional TSDF averaging. On the right, the voxelised scene.

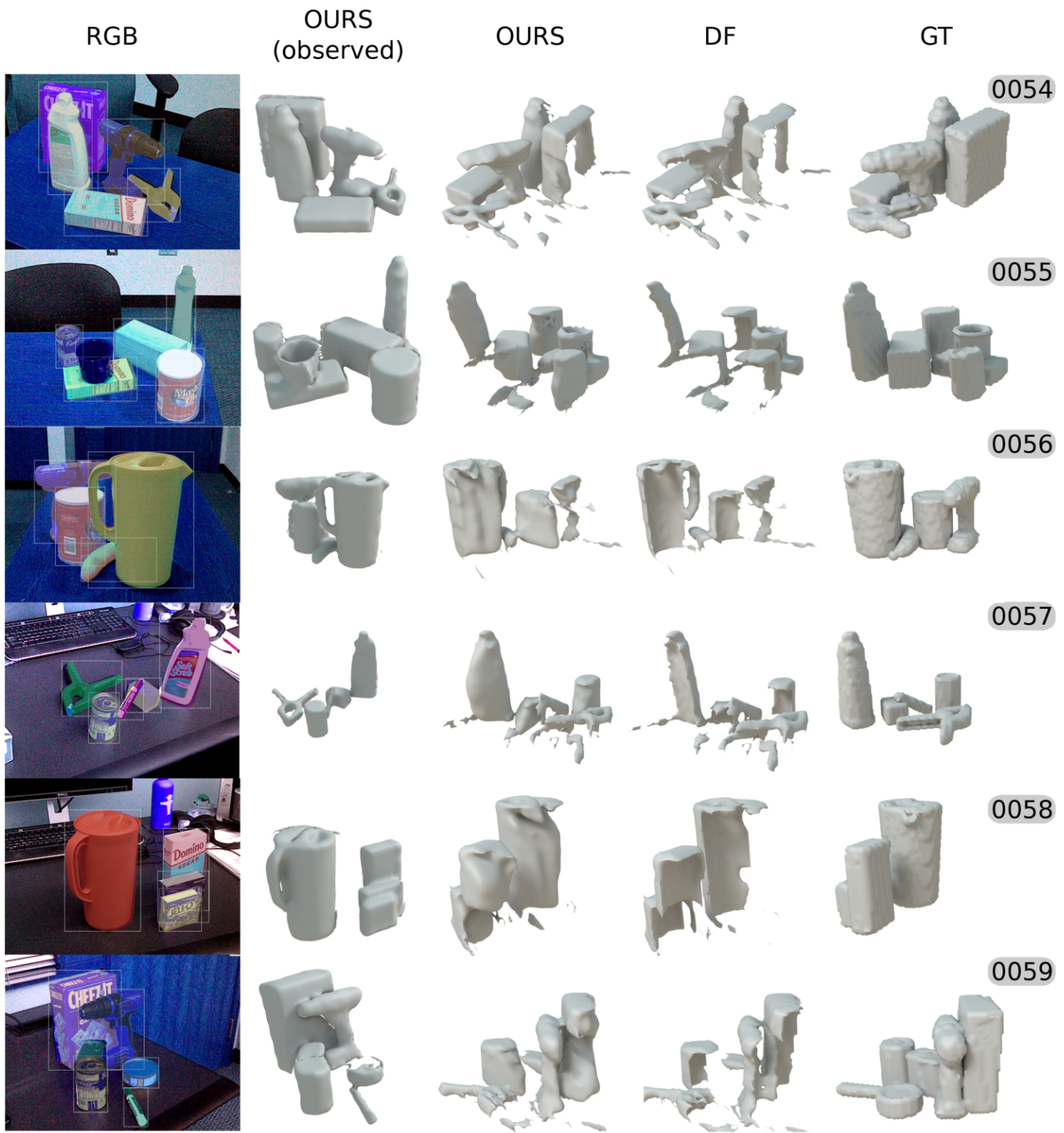

Figure 4: Reconstruction resulting from the fusion of the first 50 frames of the sequences. In this image are reported the second half of the sequences, *0054* to *0059*. From left, first RGB frame of the sequences, our reconstruction using thickness first showing the observed surfaces and then the estimated occupied space. The reconstruction labelled as DF is obtained by fusing only depth frames using traditional TSDF averaging. On the right, the voxelised scene.



\end{document}